# Classifying text using machine learning models and determining conversation drift

Chaitanya Chadha[1, a)*] and Vandit Gupta[2, a)] and Deepak Gupta [2, b)] and Dr. Ashish Khanna[2, c)]

[1]*SRM University, Haryana, 39, Rajiv Gandhi Education City, Sonipat, 131029, India*
[2]*Maharaja Agrasen Institute of Technology, Plot No 1 Rohini, Plot No 1, CH Bhim Singh Nambardar Marg, Sector 22, PSP Area, Delhi, 110086, India*

*Author Emails:*
[a)]chaitanyachadha12@gmail.com*
[b)]vanditgupta22@gmail.com
[c)]deepakgupta@mait.ac.in
[d)]ashishkhanna@mait.ac.in

**Abstract.** Text classification helps analyse texts for semantic meaning and relevance, by mapping the words against this hierarchy. An analysis of various types of texts is invaluable to understanding both their semantic meaning, as well as their relevance. Text classification is a method of categorising documents. It combines computer text classification and natural language processing to analyse text in aggregate. This method provides a descriptive categorization of the text, with features like content type, object field, lexical characteristics, and style traits. In this research, the authors aim to use natural language feature extraction methods in machine learning which are then used to train some of the basic machine learning models like Naive Bayes, Logistic Regression, and Support Vector Machine. These models are used to detect when a teacher must get involved in a discussion when the lines go off-topic.

## INTRODUCTION

Being in a team of working professionals we at times find ourselves in meetings where many times the conversation at hand goes off-topic which eventually leads to a waste of time. In schools and universities, a similar phenomenon is observed. The motivation behind our study is to try and detect when a conversation goes off-topic so the mediator or lecturer can bring the conversation back on track to avoid waste of time.

For the course assignment in the *Natural language processing* class, we decided to tackle a text classification problem on the IMapBook dataset. The dataset contained short discussions between primary school students that were chatting about different book topics. Each record was annotated with 16 attributes. Original messages were posted in the Slovene language, but they were also translated into English. We also had the information about the topic they were discussing, if their message was an answer to some previously asked question and if their discussion was relevant to the topic since there were no constraints so they could write anything they wanted. If the discussion was moving away from the proposed topic, the teacher should intervene and guide it back by asking some questions relevant to the book. The dataset contained approximately 3500 messages about 3 different short stories. Our goal was to develop models which could detect the topic of the current debate so the teacher could intervene. The idea was to develop different models and combine their outputs. Our first idea was to analyse separate messages and define their relevance to the topic. We would also have to define the type of message which could either be an answer or a question along with the message category. By combining the results of separate messages, we could determine when the conversation started to move away from the topic [8].

The main contributions of this paper are
- A proposed mechanism to determine when a conversation is drifting away from the topic.
- This research can be repurposed for other languages and productized to create web plugins that can help to detect when a conversation goes off-topic in real-time.

## RELATED WORK

Text classification is a popular topic in the natural language processing (NLP) field, thus there are many researchers working on it. In this section, we present the most relevant work and techniques they used.

Most of the research work and the state-of-the-art results were achieved using the English language, but some of the techniques and approaches can be adapted to the Slovene language. The authors of paper [1] had a similar problem, where they analysed Spanish tweets which are also relatively short texts. They discuss different approaches for pre-processing data to extract the most relevant features which are later used for classification [10]. Few standard approaches are discussed like how to define a basic term that is used by classification algorithms. Such as uni-grams (1-grams), bi-grams (2-grams), tri-grams (3-grams), n-grams. They found out that having n larger than 3 does not improve results [11]. They also tried combining different types of n-grams (like unigrams and bi-grams) to achieve better results. That also means that the attribute list was larger so they removed some entries by setting a threshold value. With threshold, they removed n-grams that did not appear frequently enough or appeared too many times since they were considered noise. They also discuss how important stemming and lemmatization are compared to the English and Spanish languages which are also interesting for the Slovene language which is also morphologically richer than English [2].

In the fourth paper [4] authors presented the recurrent convolutional neural network for text classification. The network captured contextual information of the sentence, extracted features, and learned some word representation. As described, the disadvantage of the recurrent network is that although the context of the word is captured, the model is biased where later words are more dominant than earlier words. To tackle this problem, they applied additional convolutional and pooling layers. They learned a word representation and used it for some text classification.
Another similar approach is used in the third paper [3] where the authors used convolutional neural networks for text categorization where the word order was also taken into account. The input to the network is not a standard bag-of-word representation but they present their own word representations which are some higher dimensional vectors where 2D convolution is applied. For the baseline model, they used a support vector machine (SVM) classifier with bag-of-word representation and showed that their approach gave them a lower error rate than the standard models.
There are also many already pre-trained word representations that can be used, also for the Slovenian language. The word embeddings induced from a large collection of Slovene texts composed of existing corpora of Slovene were prepared and published on CLARIN [6]. This could also be useful with our task since the embeddings were learned on bigger corpora than are available to us.

The results of the previous similar works are presented in a tabular form below.

**TABLE 1.** Comparative analysis of similar research work

| Research Paper | Model Used | Best Results |
|---|---|---|
| Antonio Fernández et al. [1] (2013) | NaiveBayesMultinomial | Accuracy: 42.38 % |
| Rie Johnson et. al [3] (2015) | CNN | Error Rate: 7.67 % |

# INITIAL IDEAS

To determine if the teacher must intervene, we need to answer the following questions:

- Are the message books relevant?
- What type is the message?
- In what category does it belong?

Based on this information we could then determine if the conservation is in need of an intervention or not.

Because there are three separate requirements our first idea was to come up with three separate classifiers. We would start with standard text classification procedures like tokenizing, stemming, and removal of stop words and then represent words as vectors in order to use them in our machine learning algorithms. After that, we would probably have to use some kind of machine-learning approach [9]. Recurring convolutional neural networks or SVMs could be used to make use of the sequential information of words as well.

## Book Relevance

To determine if the teacher must intervene, we need to answer the following questions:
- Category of the message is a good indication of whether the message is book relevant. So, if the message is classified as having a category discussion it is a good chance that the message is book relevant. So, the result of the message category classifier could be used here to determine if the book is relevant.
- Conversations have some retention. If the conversation starts leaning towards a discussion of a book most messages will be about the book, and if the conversation starts to move towards some other category most of the messages will follow. So here the sequence and previous states could be deemed important.

A good idea would be to include the original texts from the three books that our gathered messages are referring to. Another possible approach to classify book relevance would be to use the result of the message category (see section Message category) in order to get better results.

## Type of the message

Here we try to determine the type of the message. It can be a statement, a question, or an answer. Here too we drew some conclusions from the data available:
- Answers tend to follow questions. So, message order is important.
- Answers are mostly regarded as book-relevant and statements are not.

## Message Category

Each message can be one of the following:
- **Chatting,** which doesn't fall into any of the below categories;
- **Switching,** which mostly consists of asking for help;
- **Discussion,** which consists of some particular keywords ('lakho', 'bi') and descriptions of activities, objects;
- **Moderating,** where the teacher leads the conservation and which we could identify by maybe checking for grammatically correct sentences;
- **Identity,** where we could check the appearance of question marks or question words;
- **Other,** where there is mostly gibberish.

Some manual features could be added which would be especially efficient for determining the categories **other** and **identity**. Very long words or those that contain a single repeating character can quickly be classified as" other" as well as those that only contain emojis or other special characters. Detecting question marks, question words, and possibly personal names can largely contribute to classifying messages as being of type identity.

# METHODS

The steps used in the paper are represented using a flowchart in Figure 1.

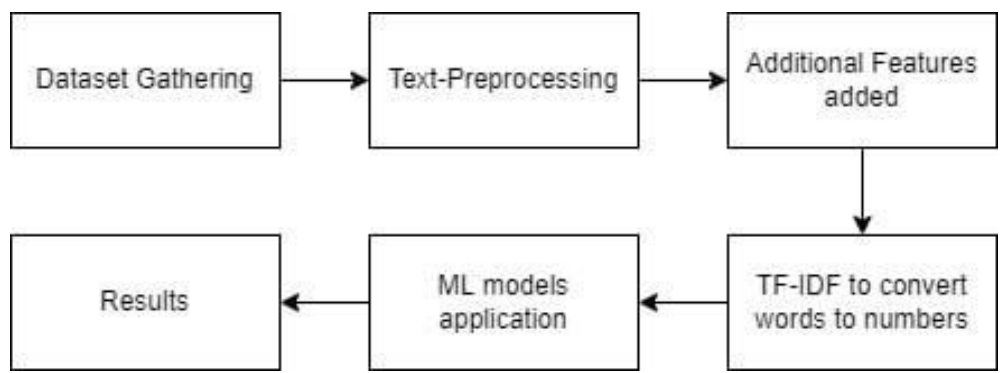

**FIGURE 1.** The working flow of the model

All the steps mentioned in the flowchart are explained in detail below

The IMapBook dataset has been used for research in this paper.

In this section, we present the pre-processing steps we performed and all the experiments we performed.

## Text pre-processing

We followed some standard text pre-processing steps. First, we tokenized each chat. After that, we used a lemmatizer to get the basic forms of each token. Both of the steps were first performed using the standard NLP tools which are available mostly for the English language. Since the Slovene language is a bit different the tokenization and lemmatization were not always correct. For that reason, we used a tool that was designed and trained for the Slovenian language [5]. Using it we obtained a better token representation of our texts which were later also lemmatized using the same tool. Lemmatized tokens were then used to build different vector representations (e.g., TF-IDF) which were used to train our models. We also found a stopword dictionary for the Slovene language but when removing the stop words the performance in all models dropped. We assume that the reason is that most of the texts are very short and thus after removing the stop words we end up with an even smaller set of words that cannot be used to successfully train the classifier.

## Additional features

With the help of the FeatureUnion class in the Sklearn library we added our own feature extractor that was combined with a Tf-idf Vectorizer. We first added all of the relevant stories to the training set, from which we removed all of the special characters which greatly improved our results. After that, we checked the messages and their appropriate tags and came to some conclusions and predictions that we predicted to be of some value. We then checked how these features improved the classification results and retained only those that provided good results. All of the additional features can be seen in Table 2.

**TABLE 2.** Manually extracted features

| Classification | Custom Features |
| --- | --- |
| Book Relevance | Length of the word |
|  | Longest repetitiveness of a word |
| Type | Contains question words or a question mark |

| Category | Length of the word |
|---|---|
| | Longest repetitiveness of a word |
| | Contains discussion words |
| | Contains identity words |

*Book relevance*

For book relevance, we checked for some characteristics that mainly separated chat messages from book relevant ones. We added three features that were observed to be important when determining book relevance. When children talked about a book, they changed their style of writing so that certain words started to appear and they avoided writing gibberish.

The first idea was to check the length of the longest word. As Slovene does not have really long words, we determined that the length of the word that was over 12 characters in length should be noted.

Next was the number of repentances of a character. Again, in Slovene, this doesn't appear often but it does in the dataset when the messages are either gibberish or the conversation has moved to a more relaxed level and with that generally isn't book relevant.

The last feature was to check for the presence of the word 'lahko' which generally referred to book-relevant messages. Because of the nature of the questions presented to children most of their sentences included the word 'lahko'. This however should not be included if the style of the questions presented to the children would change. So, this last feature is only useful if the questions retain the same structure.

*Type*

To check the type of the question we only added one feature that seemed to give good results and that was to include whether any question marks or question words appear in the sentence. To determine answers from statements we would have to take into account the order of messages and couldn't find any features that could differentiate between the two.

*Broad Category*

Here we included the first two features already mentioned in the subsection 'Book Relevance'. Like before the longest repentance and length of the word seemed to provide good indications that the message is either about chatting or other.

Another feature was for identity. Here the keywords 'kdo', 'jaz', and 'ime' seemed to mainly appear in. This can however be useful without overfitting to training data as these words are often used in identification scenarios. Checking for personal names and person's nicknames could also provide an improvement when trying to identify this category.

The last feature that we added was to check for the presence of words 'lahko' and 'bi', which often indicated that the message was of the type of discussion. Again, as already mentioned in the subsection 'Book Relevance' this feature should also be taken with a grain of salt as it does help with the identification of discussion-related messages but only because the questions posed to the children were formatted to provide answers or messages with these words.

# RESULTS

## Models

Initial goal was to detect when the teacher needs to intervene in the discussion. To achieve it we defined several classification models. The first set of models was developed to classify each chat into one of the two classes. Either the chat is somehow relevant to the book or not. The other set of models was more complicated. They were classifying the chats into 6 groups. Each group represents a simple description of what the text is talking about. All the categories are described in a separate document.

The models we used were Naive Bayes, Logistic Regression, and Support Vector Machine (SVM). Each model was trained on 2653 random examples and tested on the rest of 1062. The distribution of relevant and non-relevant text is plotted in Figure 2 for the training set (left) and testing set (right). We evaluated them by calculating the accuracy, recall, precision, and F1 score. The majority classification accuracy of predicting the relevance into the non-relevant class was 0.619.

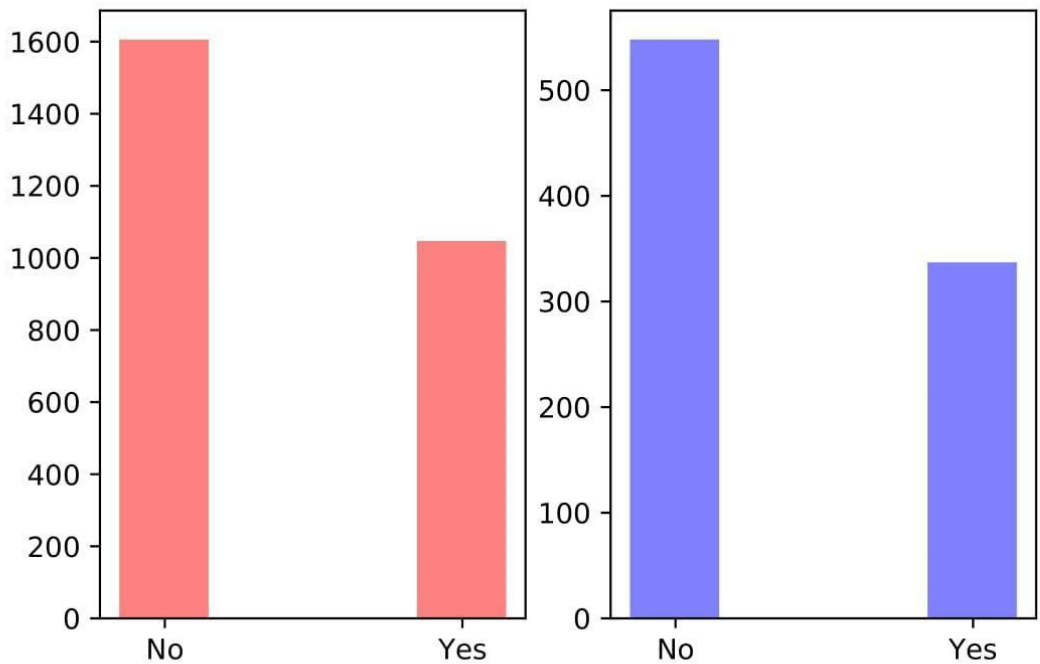

**FIGURE 2.** Distribution of relevance on train and test set

All the results from the models we tried are shown in Table 3. For each of the models, the first line represents the score when classifying the 'Book relevance'. The second row is the score for predicting the 'Category' while the third one is the score for classifying 'Type'.

*Naive Bayes*

The simplest model that we tried was Naive Bayes which assumes that words in the sentence are independent.

**TABLE 3.** Evaluation of models

| Models | Accuracy | Precision | Recall | F1 |
|---|---|---|---|---|
| **NB** | 0.81 | 0.80 | 0.66 | 0.73 |
| | 0.50 | 0.70 | 0.50 | 0.56 |
| | 0.46 | 0.64 | 0.46 | 0.49 |
| LR | 0.85 | 0.84 | 0.75 | 0.80 |
| | 0.73 | 0.73 | 0.73 | 0.72 |
| | 0.76 | 0.77 | 0.77 | 0.76 |
| SVM | 0.86 | 0.84 | 0.79 | **0.81** |
| | 0.74 | 0.75 | 0.74 | **0.73** |
| | 0.78 | 0.78 | 0.78 | **0.78** |

Although the model is naive, the results we got are not that bad. The AUC of the model for predicting the relevance of the text is 91%.

*Logistic Regression*

The performance is similar to Naive Bayes. But we got an AUC of 92% which is the highest of all baseline models [7].

*Support Vector Machine*

Overall SVM performed the best so we will use it in further experiments. We also plotted the ROC curve for predicting the relevance of the text which is shown in Figure 3.

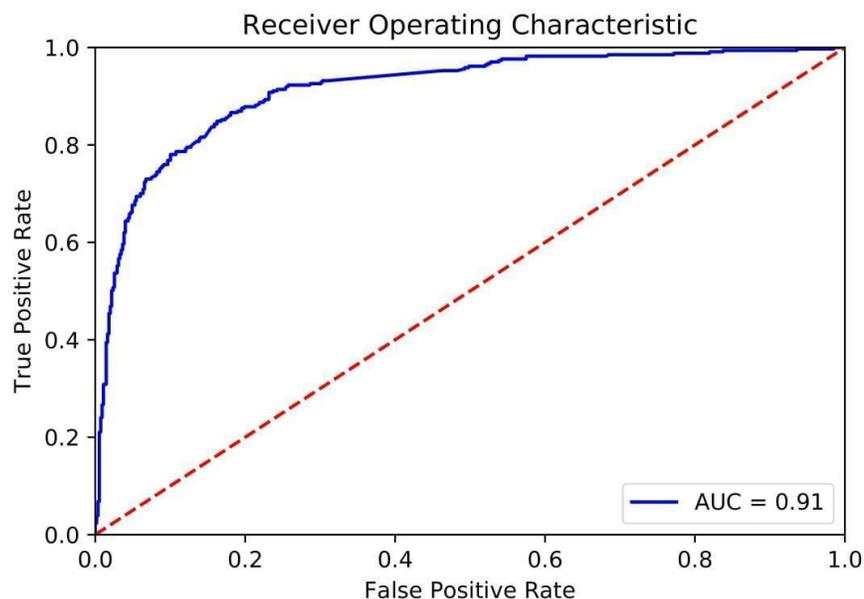

**FIGURE 3.** ROC curve of SVM performance

To demonstrate the performance of the SVM model when classifying the Category, we plotted two confusion matrices where we can observe the distribution of the testing set into the different categories and how our SVM model predicted them which is shown in Figure 4. In Figure 5 we plotted the same normalised distribution.

From Figure 4 we can observe that most of the test examples are from category C (Chatting) and we are able to correctly classify 83% of them. We can also observe that 10% of the examples are wrongly classified as category D (Discussion).

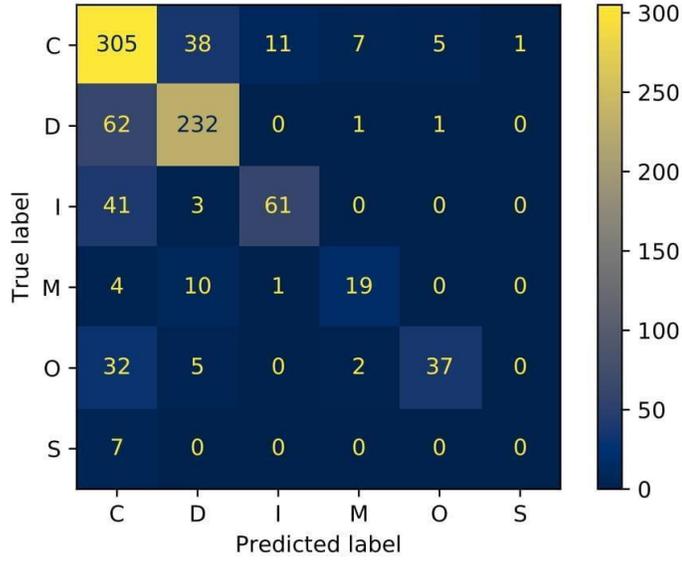

**FIGURE 4.** Confusion matrix of classifying the category

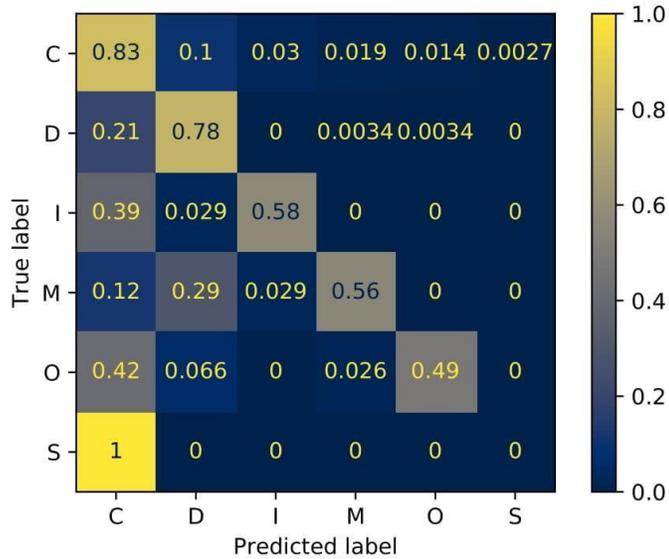

**FIGURE 5.** Normalised confusion matrix

The examples from the second category D are correctly classified in 78% of cases [10]. In 21% of cases, the model wrongly classifies it as category C. category I (Identity) is correctly classified in 58% of the cases. Most of the wrongly classified examples are again put into category C. For category M (Moderating) we can correctly predict 56% of the true examples. Almost 30% are wrongly classified into category D and another 12% into category C. In category O (Other) we correctly classified only 49% of the examples. Most of them (42%) are wrongly classified into category C. In the case of category S (Switching), we wrongly classify 100% of the examples into category C. From Figure 4 we can observe that in this case there were only 7 testing examples.

We can assume that our model correctly predicts most of the examples from category C. We are also quite sure that the examples from category D are correctly classified (with a probability of 0.78). For other categories, the probabilities are lower (around 50%) except for the Switching category where our model fails completely. We can also observe that most of the wrong predicted labels are classified into category C.

## Custom features impact

Some of the custom features for classifying messages into a particular group worked better than others. In table 4 the improvements for our classification models can be seen with the included custom features.

**TABLE 4.** Performance Improvements in F1 score with used custom features.

| Classification | Model | Improvement in F1 |
|---|---|---|
| **Book Relevance** | NB | +1,3% |
| | LR | -0,2% |
| | SVM | +0,1% |
| **Type** | NB | -0,2% |
| | LR | +0,3% |
| | SVM | +0,2% |
| **Category** | NB | -0,5% |
| | LR | +1,2% |
| | SVM | +1% |

As it can be seen we achieved the best results when using the custom features when trying to classify messages based on the broad category. There, our score was improved by a little more than 1% for the LR and SVM models. This was not expected as well-defined features for the type classification problem were expected to provide the best results. There, an improvement of less than 0.5% was observed and after further investigation, we identified that the problem wasn't so much in wrongly classifying questions as it was in wrongly classifying answers and statements.
Using custom features for classifying book relevance did show an improvement in the Naive Bayes F1 score by 1.3%. When tackling this problem, we saw a much greater improvement in our F1 score when we included the provided stories in the training data as defined in the subsection" Additional features", as we did with our manual features. Even though the words 'lahko', 'bi', and 'da' often appeared in messages that were labelled as book relevant they were most probably already detected as of great value to determine the type by our models and so didn't provide much better results.

## Detection of conservation drift

To detect when the teacher needs to intervene, we trained two models. They were trained using the lemmatized texts as before. But in this case, we took sequential texts to train and test our models. For training, we took the first 70% of lemmatized texts and calculated their Tf-idf vectors using our custom features. From the experiments above we decided to train the SVM model for the classification of the relevance of the text and Linear Regression (LR) for the classification of categories. Both of the models were trained and evaluated using the 5-fold cross-validation. The accuracy and F1 score of the SVM model were 0.83 and 0.82 respectively. For the LR model, we obtained an accuracy of 0.65 and an F1 score of 0.67.

The last 30% of conversations we used to detect the drift. The idea was to take batches of sequential messages and for each, we predicted the relevance label. We counted the number of relevant chats in a batch using the following method. If the label was positive ('Yes') we counted it as relevant. If the label was negative ('No') we also predicted the category. We defined some soft categories which also count as relevant. If the category was any of those, we classified chat as relevant otherwise as non-relevant. From the frequency of relevant/non-relevant chats, we classified each batch. The drift was detected if there were non-relevant discussions in a few consecutive batches.

To determine the number of sequential messages and the number of consecutive batches we manually extracted some features from the training set. First, we calculated how many sequential messages are non-relevant before the teacher had to intervene in the conversation. We averaged the results over the whole training set and got an average number of 4.8. This means that when there was no relevant discussion between the participants the teacher on average intervened after 5 non-relevant messages. We were also interested in how often are relevant messages and how many non-relevant chats are between two relevant messages. The number we got was 2.5 but since we are not interested in the consecutive relevant messages, we corrected this number. We defined that if we observe more than 3 sequential relevant messages, we conclude that this is a relevant conversation and thus there are no non-relevant messages in between. The corrected average we got was 4.0. Meaning that during two relevant messages (which are not marked as relevant conversation) there are on average 4 non-relevant messages. Since we are also only 78% sure that when we predict the category D it is the correct label, we increased both of these numbers by 50%. We defined the size of the batch as the average number of messages before the teacher intervenes which is increased by 50% and thus, we get 6. The number of consecutive batches is defined from the corrected average of non-relevant chats between relevant ones. By also increasing it by 50% we get 7 consecutive batches. Since 30% of messages from category M are also wrongly classified as category D, we define two soft categories M and D which are both seen as relevant.

We notify the teacher that the conversation moved away from the topic if there is no single relevant message in the 7 consecutive batches of 6 sequential messages.

## DISCUSSION

We achieved quite good results using the traditional natural language processing approaches. We mainly focused on implementing and refining these and have gotten satisfying results. Our best results were with the use of the Support Vector Machine classifier equipped with custom features that gave us F1 scores of 0.81, 0.73, and 0.78 for the classification task of determining book relevance, category, and type of message respectively. Our implementation also gave us good results when tackling the main problem of detecting conversation drift. We trained two models using our manually extracted features. Here we had to determine some additional parameters. By manually inspecting the dataset and calculating some specific properties we were able to correctly determine the sizes of batches and the number of messages in the batch. We found out that the results very much depend on how well we are familiar with the data we use and how many distinguishing features we can extract. There may be better approaches but even with this one, we were able to quite successfully and reliably detect when the conversation started to move away from the original discussion so we could inform the teacher about it.

We also started experimenting with more advanced models where features are learned and we do not need to manually extract them. Since we were limited by time and hardware the results were not comparable to the ones, we got by manually extracting features.

## CONCLUSION

We were successful in implementing various classifiers for determining book relevance, category, and type of messages, and by combining them we built a good detector for conversation drift. Although we could have gotten better results using more advanced methods and models, we rather chose to focus our efforts on improving the results using more traditional machine learning models and manually extracting different features depending on the problem we were solving. Future directions are to train some deep neural models. Manual feature extraction takes a long time and is not as abstract as it could be. These manual features contributed to improving the category classification but didn't perform as well for the type or book relevance classification. For the following, more work would have to be done to come up with better features. We also need to be very familiar with our data and thus we design features that are consistent with our dataset which is another bias of our model. With deep learning approaches we could build a network that would be able to learn this feature on its own and they would be less biased than the current ones. In the end, we are successfully able to determine the conversation drift in a conversation with high accuracy and this research can be further used in various fields.

# REFERENCES

1. Antonio Fernández Anta, Philippe Morere, Luis Chiroque, and Agustín Santos, "Sentiment analysis and topic detection of Spanish tweets: A comparative study of NLP techniques" in *Procesamiento de Lenguaje Natural*, Volume 50, 2013, pp. 45–52
2. Essam H Houssein, Zainab Abohashima, Mohamed Elhoseny, and Waleed M Mohamed, "Machine learning in the quantum realm: The state-of-the-art, challenges, and future vision" in *Expert Systems with Applications*, Volume 194, 2022, Article. 116512
3. Rie Johnson and Tong Zhang, "Effective Use of Word Order for Text Categorization with Convolutional Neural Networks." in *NAACL,* 2015
4. Siwei Lai, Liheng Xu, Kang Liu, and Jun Zhao, "Recurrent convolutional neural networks for text classification" in *Twenty-ninth AAAI conference on artificial intelligence (AAAI'15). AAAI Press, pp.* 2267–2273.
5. Nikola Ljubešić, "The CLASSLA-StanfordNLP model for lemmatization of non-standard Slovenian 1.1" in *Slovenian language resource repository CLARIN.SI*, 2020.
6. Nikola Ljubešić, Tomaž Erjavec, and Darja Fišer, "Adapting a State-of-the-Art Tagger for South Slavic Languages to Non-Standard Text" in *Proceedings of the 6th Workshop on Balto-Slavic Natural Language Processing,* 2017, pp. 60–68, Valencia, Spain. Association for Computational Linguistics.
7. Bernhard Schölkopf, "Causality for machine learning" in *Probabilistic and Causal Inference: The Works of Judea Pearl*, 2022, pp. 765–804.
8. Ricardo Vinuesa and Steven L Brunton, "Enhancing computational fluid dynamics with machine learning" in *Nature Computational Science*, Volume 2, Issue 6, 2022, pp. 358–366.
9. Qi Wang, Yue Ma, Kun Zhao, and Yingjie Tian, "A comprehensive survey of loss functions in machine learning." in *Annals of Data Science*, Volume 9, Issue 2, 2022, pp. 187–212.
10. Sean Whalen, Jacob Schreiber, William S Noble, and Katherine Pollard, "Navigating the pitfalls of applying machine learning in genomics." in *Nature Reviews Genetics*, Volume 23, Issue 3, 2022, pp. 169–181.
11. Lorijn Zaadnoordijk, Tarek R. Besold, and Rhodri Cusack, "Five Lessons from infant learning for unsupervised machine learning" in *Nature Machine Intelligence*, Volume 4, Issue 6, 2022, pp. 510–520.